# On Ternary Coding and Three-Valued Logic

Subhash Kak

**Abstract.**
Mathematically, ternary coding is more efficient than binary coding. It is little used in computation because technology for binary processing is already established and the implementation of ternary coding is more complicated, but remains relevant in algorithms that use decision trees and in communications. In this paper we present a new comparison of binary and ternary coding and their relative efficiencies are computed both for number representation and decision trees. The implications of our inability to use optimal representation through mathematics or logic are examined. Apart from considerations of representation efficiency, ternary coding appears preferable to binary coding in classification of many real-world problems of artificial intelligence (AI) and medicine. We examine the problem of identifying appropriate three classes for domain-specific applications.

**Introduction**
The problem of optimal coding of numbers has been examined by many scholars (e.g. [1][2]) and it is known that ternary coding is more efficient than binary coding [3]. The reason why it is not used extensively in hardware is because technology for binary processing is already established and the implementation of ternary coding is more complicated in computation, although it is used in digital communications where error correction coding is employed [4]. Ternary coding remains relevant in algorithms that use decision trees.

Let's consider coding of numbers to an arbitrary base b. Given that the probability of the use of the b symbols can be taken to be the same and equal to $1/b$, the information associated with each symbol is $\log b$. The efficiency of the coding scheme per symbol is

$$E(b) = \frac{\ln b}{b} \qquad (1)$$

To find the value of *b* for which it is a maximum we take its derivative with respect to *b* and equate that to zero. This yields the condition that $\ln b = 1$, from which we conclude that the optimal base is *e*, with *E*(e)= 0.368 nats or 0.531 bits.

Although *e* cannot be used for a rational representation of numbers, we can use it as a measure to mark the performance of other values of *b* as in Table 1.

Table 1. Efficiency of coding for certain bases

| b | 2 | e | 3 | 4 | 5 | 8 | 10 | 12 | 20 | 100 |
|---|---|---|---|---|---|---|---|---|---|---|
| E(b) nats | 0.347 | 0.368 | 0.366 | 0.347 | 0.322 | 0.260 | 0.230 | 0.207 | 0.150 | 0.046 |
| E(b) bits | 0.500 | 0.531 | 0.528 | 0.500 | 0.465 | 0.375 | 0.331 | 0.298 | 0.216 | 0.066 |

The efficiency is quite close to the maximum for b=3 (a value of 0.366 as compared to the optimal value for e which is 0.368), with the next best value coming at the bases 2 and 4. The efficiency at b=3 is off from the optimal by 0.5 percent, whereas it is off by 5.7% at b=2. After this the values decline monotonically as shown in Figure 1 in which the most interesting region is for the base values that lie between 2 and 4.

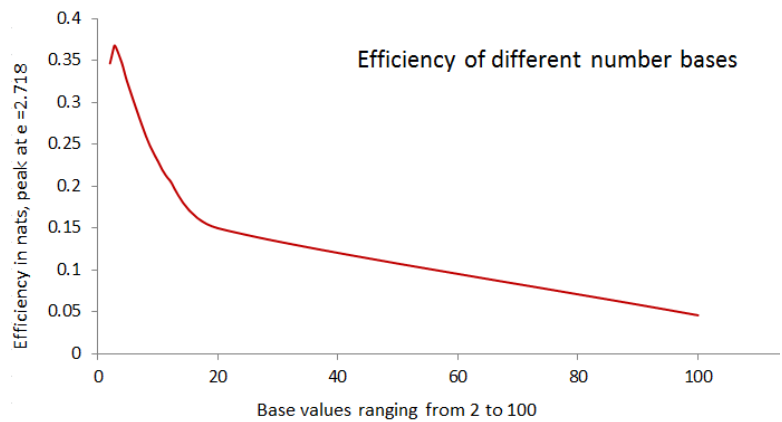

Figure 1. Efficiency of different number bases

If we consider only the capacity of a code to carry information, then binary comes up considerably short compared to the optimal value. For binary, the value is $\ln 2 = 0.693$ which is short by $\ln \frac{e}{2} = 0.307$ from the optimal. (In bits, the corresponding numbers are 1.0 and 0.443, respectively.) For ternary, the coding is vastly more efficient since $\ln 3 = 1.099$ and $\ln \frac{e}{3} = -0.099$. As we see, the carrying capacity differences are much more significant compared to those based on efficiency measure given in equation (1). For example, for b = 2, the carrying capacity is less by 30.7%.

In this paper we present a new comparison of binary and ternary coding and their relative efficiencies are computed both for number representation and decision trees. The implications of our inability to use optimal representation through mathematics or logic are investigated. From a practical point of view, we note the importance of the use of ternary logic in medicine and artificial intelligence (AI) and the expectation that it will lead to better performance in certain applications.

**Philosophical issues**
Since coding has aspects that go beyond representation of data, the issue of optimal coding leads to many philosophical questions. Since binary is not optimal, there ought to be a logic that has more than two classes but it is not clear how such a logic manifests itself in natural processes or in cognition.

Absolute optimality appears to be unachievable as it corresponds to a non-integer, irrational value. We may conclude that mathematics is not an optimal or complete way to represent reality and this shortcoming may be added to the known limitations of mathematics [5].

Although interactions with Nature are normally seen as responses to binary questions in terms of "yes" or "no" [6][7], this is incorrect if a third category of "maybe" is included.

This third "indeterminate" category may be seen as the origin of non-classical aspects of reality. A relevant parallel from quantum mechanics is that the observation cannot reveal what the state of the system was prior to this observation [8][9]. This indeterminacy creates unique challenges if one wishes to build a quantum system for computation [10][11].

Sometimes physicists search for mathematical arguments for the ontological basis of reality and why one has symmetries of different kind (as in [12]). Speaking in a similar vein, one may say that the universe is 3-dimensional for that gives the optimal coding in integer values. Likewise, one may add that the brain performs a 2.5 perception [13] for it makes us come close to the optimal value on the "screen" of our mind.

**Comparison for bases 2 and 3**

The two bases of most interest are 2 and 3. Consider the count of all numbers over the range $(0, N = 2^n - 1)$, and let the number of bits we come across when the numbers are written in the most compact form (for example, 001 is written as 1, and so on) be represented by S. We will use the product $R = S \times b = Sb$ to be the range associated with the representation so that this range can be meaningfully compared across different bases.

In general, the value of S to any base is as follows:

$$S_b(N = b^n - 1) = \sum_{j=1}^{j=n} j(b^j - b^{j-1}) \qquad (2)$$

This can be simplified as given below:

$$S_b(N = b^n - 1) = \frac{1}{b-1}[n(b^{n+1} - b^n) - b^n + 1] \qquad (3)$$

For the special case of b=2, this simplifies to:

$$S_2(N = 2^n - 1) = (n-1)2^n + 1 \qquad (4)$$

Thus when n=3, that is N=7, we get the numbers 1,10,11,100,101,110,111 for a total of 17 locations that equals 17 bits. The formula (2) gives us $2 \times 2^3 + 1 = 17$, and the corresponding $S_2 2$ is 34.

For N=8, for which n=2, the numbers are 1, 2, 10, 11, 12, 20, 21, 22 for a total of 14 symbols which is the same $S_3(8) = 1/2[2(3^3 - 3^2) - 3^2 + 1]$. The $S_3 b$ count of the range is $14 \times 3 = 42$.

For a comparison for N=8 for binary (b=2) coding we get to add 8 more (for 8=1000) to the previous count of 34 to get the same range count of 42 as for base 3. If we throw in b=4 for another point of comparison, the numbers are 1, 2, 3, 10, 11, 12, 13 for $S_4(7)=11$ and the corresponding range of 44.

Table 2 provides the total range count for the bases 2, 3, and 4 for different N.

Table 2. Range counts

| N | 1 | 2 | 3 | 4 | 7 | 8 | 9 | 15 | 16 | 26 | 27 | 31 | 63 | 64 | 80 | 100 |
|---|---|---|---|---|---|---|---|---|---|---|---|---|---|---|---|---|
| Sb, b=2 | 2 | 6 | 10 | 16 | 34 | 42 | 50 | 98 | 108 | 208 | 218 | 258 | 642 | 656 | 880 | 1160 |
| Sb, b=3 | 3 | 6 | 12 | 18 | 36 | 42 | 51 | 105 | 114 | 204 | 216 | 264 | 648 | 660 | 852 | 1152 |
| Sb, b=4 | 4 | 8 | 12 | 20 | 44 | 52 | 60 | 108 | 120 | 220 | 232 | 280 | 684 | 700 | 956 | 1276 |

We observe that the best R=Nb values keep on switching between the bases 2 and 3, depending on how close one is to the transitions at the powers of these bases. At $2^i-1$, the base 2 counts are a bit lower (the differences are 2, 2, 7, 6, 6), but at $3^i-1$, the counts of base 3 are lower by a much more substantial margin (0, 4, 8); and at N=100, it remains 8.

A comparison of bases 2 and 3 on absolute count (Figure 2) shows that their difference is relatively small.

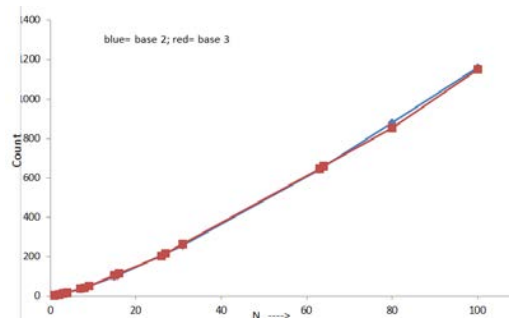

Figure 2. Comparison of bases 2 and 3

**Ternary trees**
Trees are ubiquitous in data structures [14]. Every node (excluding a root) in a tree is connected by a directed edge from exactly one other node, which is the parent node. Alhough binary trees are used most frequently, there are situations where ternary trees offer advantages.

Excellent trees for data values associated with different probabilities are given by Huffman coding or arithmetic coding [15]. Huffman coding is optimal if all symbol probabilities are integral powers of the inverse of the coding alphabet. This is hardly ever going to be true of a real-world situation and in those cases Huffman coding can take up to one extra bit per symbol.

By dispensing with the restriction that each symbol must translate into an integral number of bits, arithmetic coding achieves greater efficiency. But since its implementation is more complex, we shall limit our discussion to Huffman coding.

The coding of different symbols (or messages) of unknown probabilities by efficient search trees is equivalent to representation by sequences as shown in Figure 3 for 9 symbols.

We will first assume that the probabilities of the symbols are the same, then a modification of a straight enumeration of the symbols should provide the most efficient coding.

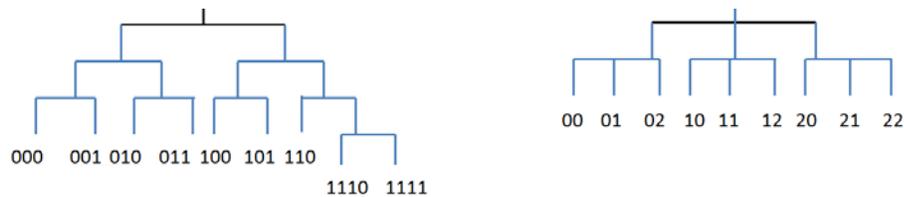

Figure 3. Binary and ternary trees

In the case of the example of Figure 3, the average length of a symbol in the binary case is 29/9=3.22 bits, whereas ternary coding shown on the right gives us 2 trits. Since 1 trit equals $log_2 3 = 1.585$ bits, ternary coding is more efficient for this example.

In the enumeration mapping, which must satisfy the prefix condition, we run through all the elements of one size and then replace the largest number by the next sized number. The counting will be as shown here for both binary and ternary (Table 3).

Table 3. Coding for different size symbols

| Binary | Ternary |
|---|---|
| 3: 0, 10, 11 | 3: 0, 1, 2 |
| 4: 00, 01, 10, 11 | 4: 0, 1, 20, 21 |
| 5: 00, 01, 10, 110, 111 | 5: 0, 1, 20, 21, 22 |
| 6: 00, 01, 100, 101, 110, 111 | 6: 0, 10, 11, 20, 21, 22 |
| 7: 00, 010, 011, 100, 101, 110, 111 | 7: 0, 10, 11, 12, 20, 21, 22 |
| 8: 000, 001, 010, 011, 100, 101, 110, 111 | 8: 00, 01, 10, 11, 12, 20, 21, 22 |
| 9: 000, 001, 010, 011, 100, 101, 110, 1110, 1111 | 9: 00, 01, 02, 10, 11, 12, 20, 21, 22 |

A comparison between binary and ternary search trees for unknown symbol frequencies is given in Table 4.

Table 4. Comparison of binary and ternary coding for different number of symbols

| Symbols | 3 | 4 | 5 | 6 | 7 | 8 | 9 | 10 |
|---|---|---|---|---|---|---|---|---|
| Binary (bits) | 1.66 | 2 | 2.4 | 2.7 | 2.86 | 3 | 3.2 | 3.4 |
| Ternary (trits) | 1 | 1.5 | 1.6 | 1.83 | 1.86 | 2 | 2 | 2.2 |
| Ternary (bits) | 1.59 | 2.38 | 2.54 | 2.91 | 2.93 | 3.17 | 3.17 | 3.48 |

We observe that ternary is superior to binary where the number of symbols is in powers of 3; and binary is superior otherwise.

It is quite clear from the underlying combinatorics that as it is a larger set, ternary coding is more efficient than binary only for specific values as shown in Table 4. When considering non-uniform probabilities, the smaller binary code alphabet increases the advantage even further.

Now assume that the non-uniform probabilities are according to the first digit phenomenon [16], in which the probabilities are $P(d) = log_r(1 + \frac{1}{d})$, where $r$ is the base and $d= 1,… r-1$. We present this as an excurses to bring to the reader this property of natural random numbers, which might be relevant in some classification problems, although as we show below it makes no difference for efficiency of coding here.

Because of the logarithmic structure of the probabilities, there are various relations amongst them. In all of these, we have (if warranted by the presence of the relevant numbers) Prob [1] = Prob [2+3], Prob [2] = Prob [4+5], and Prob [3] = Prob [6+7]. For base 10, we also have:

 Prob [1] = Prob [5+6+7+8+9]
 Prob [2] = Prob [5+8+9]
 Prob [4] = Prob [8+9]

These values for the first digit frequencies for different bases are given in Table 5.

Table 5. First digit frequencies in different bases (3 to 10)

| Numeral | 1 | 2 | 3 | 4 | 5 | 6 | 7 | 8 | 9 |
|---|---|---|---|---|---|---|---|---|---|
| Base 3: | .631 | .369 | | | | | | | |
| Base 4: | .500 | .292 | .207 | | | | | | |
| Base 5: | .431 | .252 | .179 | .139 | | | | | |
| Base 6: | .387 | .226 | .161 | .125 | .102 | | | | |
| Base 7: | .356 | .208 | .148 | .115 | .094 | .079 | | | |
| Base 8: | .333 | .195 | .138 | .107 | .088 | .074 | .064 | | |
| Base 9: | .315 | .185 | .131 | .102 | .083 | .070 | .061 | .054 | |
| Base 10: | .301 | .176 | .125 | .097 | .079 | .067 | .058 | .051 | .046 |

We see that if we are considering first digits to any base, binary coding is more efficient, as is to be expected. We mention it here to make a distinction between the mathematical coding process and the more fundamental question of using a ternary classification of data obtained from a natural process. If Nature uses the most efficient ternary coding, it perhaps gets manifested at a level that is deeper than the use of a ternary alphabet.

**Ternary classification**

It is possible that ternary classification, when optimized appropriately, will reveal the deeper ontological basis corresponding with ternary coding but that can only be a research program at this stage. Meanwhile, let me list a few applications where ternary classification has been used. In urban analysis, a ternary representation with the three classes as cyber (C), physical (P), and human (H) gives better results [17].

Ternary classification has been used with the classes positive, negative, and neutral in many problems related to response to drugs. In the problem of classifying chemicals with biomedical properties, one may be considering agonistic, antagonistic, or no receptor activities [18].

Real world applications are associated with multi-class classification problems with unknown categories. This is true, for example, in automatic webpage annotation, and disease diagnosis with unknown classes [19]. The current regime of two testing classes (those being administered the drug and those in the control sample who are given the placebo) could be replaced by three classes (related to psychological and physical profiles together with the control class) and this should help find hitherto undiscovered correlations between mind states and health. One could also classify subjects in 3 categories, and it is significant that a ternary classification of healthy patients is used in at least one traditional medicine system [20][21].

The search for best three-class representation of data in medicine is a question that requires much analysis and research. Unsupervised clustering algorithms, such as hierarchical clustering, model-based clustering, k-means clustering, and self-ordering maps, may be used [22][23][24] and they offer improvement of performance [25]. But they are dependent on the quality and the nature of the data and they do not automatically obtain the best three-class model or reveal a fundamental ternary ontological structure.

**Ternary logics**

Three-valued logics have truth values indicating true, false, and an indeterminate third value. One can directly see the application of this to databases where some data entries are missing. Indeed, the database structural query language SQL implements ternary logic to handle comparisons with NULL field content, which stands for the missing data. This may be necessary even when an actual value exists, but it is not recorded in the database [26].

SQL uses AND, OR, and NOT tables where the intermediate value is intended to be interpreted as UNKNOWN. Explicit comparisons with NULL, including that of another NULL yields UNKNOWN.

If the truth values are represented numerically as -1, 0, 1 for false, unknown, and true, respectively, then we can set up tables for the $NOT(A)$, $A \cap B$, and $A \cup B$ where $A \cup B$ is $MAX(A, B)$ and $A \cap B$ is $MIN(A, B)$ (Table 6).

| A | NOT A |
|---|---|
| -1 | +1 |
| 0 | 0 |
| +1 | -1 |

| A∪B | | B | | |
|---|---|---|---|---|
| | | -1 | 0 | +1 |
| | -1 | -1 | 0 | +1 |
| A | 0 | 0 | 0 | +1 |
| | +1 | +1 | +1 | +1 |

| A∩B | | B | | |
|---|---|---|---|---|
| | | -1 | 0 | +1 |
| | -1 | -1 | -1 | -1 |
| A | 0 | -1 | 0 | 0 |
| | +1 | -1 | 0 | +1 |

Table 6. NOT, OR (max function), and AND (min function) tables for a three-valued logic

These functions can be used for developing different logical operations on the variables.

A similar truth table can be used to consider the specific states of photon polarization, A, that are used in quantum computations: horizontal (H = -1), entangled or diagonal (D = 0) and vertical (V = 1). The second variable, B, is the state of the measurement apparatus or filter which can be aligned to horizontal (H = -1), diagonal (D = 0) or vertical (V = +1) orientations. The groups of photons, all in the same state, is measured correctly if the output intensity is the same as the input intensity (the two coincide) or if it is zero (the output is the complement of the input); the output state is 0, if the output intensity is half of the input intensity [11][27]. This sets up the truth table of Table 8:

|   |   | B | | |
|---|---|---|---|---|
|   | A∪B | -1 | 0 | +1 |
| A | -1 | -1 | 0 | +1 |
|   | 0 | 0 | 1 | 0 |
|   | +1 | +1 | 0 | +1 |

Table 8. Truth table for a quantum state measurement problem

**Conclusions**

In this paper we presented a new comparison of binary and ternary coding, and their relative efficiencies were computed both for number representation and decision trees. The implications of our inability to use optimal representation through mathematics or logic were examined. We proposed that interactions with Nature, which are seen to be binary at the deepest level, need to be enhanced by a third category of "maybe", and this additional category is like the fundamental indeterminacy of quantum mechanics. This approach seems to be another way of looking at the classical-quantum divide, which is normally viewed through the lens of differently operating probability theory [28][29].

The most important question raised by our consideration of the near-optimality of ternary classification, and superiority over binary, is how these classes should be identified in any application. The objective is not to go to unconstrained unsupervised learning with its many classes, but rather to identify a triplicate-logic structure that offers superior understanding compared to binary classification and provides better performance in AI applications.